\documentclass[letterpaper, 10 pt, conference]{ieeeconf}  % Comment this line out if you need a4paper

\IEEEoverridecommandlockouts                              % This command is only needed if
\usepackage{graphics} % for pdf, bitmapped graphics files
\usepackage{epsfig} % for postscript graphics files
\usepackage{mathptmx} % assumes new font selection scheme installed
\usepackage{times} % assumes new font selection scheme installed
\usepackage{amsmath} % assumes amsmath package installed
\usepackage{amssymb}  % assumes amsmath package installed
\usepackage{xspace}
\usepackage{paralist}

\usepackage{subfig}
\usepackage{makecell}
\usepackage{url}
\usepackage{colortbl}
\usepackage{multirow}

\newcommand{\mypartitle}[2][2.]{\vspace*{-#1 ex}~\\{\noindent {\bf #2}}}
\newcommand{\myparit}[1]{{\vspace{1mm}\noindent {\it #1}}}

\newcommand{\NFeatureChannels}[0]{\ensuremath{N_w}\xspace}
\newcommand{\OurDataset}[0]{{DIH}\xspace}
\newcommand{\NStages}[0]{\ensuremath{T}\xspace}
\newcommand{\ConfidenceMapThreshold}[0]{\ensuremath{\tau}\xspace}
\newcommand{\PCKhDistanceThresh}[0]{\ensuremath{d}\xspace}
\newcommand{\PCKhCoef}[0]{\ensuremath{\kappa}\xspace}
\newcommand{\PCKhBBGtHeight}[0]{\ensuremath{h}\xspace}

\newcommand{\RPM}[0]{RPM\xspace}
\newcommand{\TheRPM}[0]{\RPM-2S\xspace}
\title{\LARGE \bf
Real-time Convolutional Networks\\for Depth-based Human Pose Estimation
}

\def\idiapref{\textsuperscript{\textasteriskcentered}}
\def\epflref{\textsuperscript{\textdagger}}

\author{Angel Mart\'inez-Gonz\'alez\idiapref\epflref, Michael Villamizar\idiapref, Olivier Can\'evet\idiapref\ and Jean-Marc Odobez\idiapref\epflref
\thanks{$^{*}$ Idiap Research Institute, Switzerland. \{angel.martinez, michael.villamizar, olivier.canevet, odobez\}@idiap.ch }
\thanks{\textsuperscript{\textdagger} \'Ecole Polytechnique F\'ed\'erale de Lausanne (EPFL), Switzerland.}
}

\usepackage[hidelinks]{hyperref}
\usepackage{xspace}

\makeatletter
\DeclareRobustCommand\onedot{\futurelet\@let@token\@onedot}
\def\@onedot{\ifx\@let@token.\else.\null\fi\xspace}

\def\eg{{e.g}\onedot} 
\def\ie{{i.e}\onedot} 
 
\def\etc{{etc}\onedot}

\def\etal{{et al}\onedot}
\makeatother

\usepackage{flushend}

\begin{document}

\maketitle
\thispagestyle{empty}
\pagestyle{empty}

%%%%%%%%%%%%%%%%%%%%%%%%%%%%%%%%%%%%%%%%%%%%%%%%%%%%%%%%%%%%%%%%%%%%%%%%%%%%%%%%
\begin{abstract}
  We propose to combine recent Convolutional Neural Networks (CNN) models with depth imaging to obtain a reliable and fast multi-person pose estimation
  algorithm applicable to Human Robot Interaction (HRI) scenarios.
  Our hypothesis is that depth images contain less structures and are easier to process than RGB images while keeping the required information
  for human detection and pose inference, thus allowing the use of simpler networks for the task.
  Our contributions are threefold.
  (i) we propose a fast and efficient network based on residual blocks (called \RPM)
  for body landmark localization from depth images;
  (ii) we created a public dataset \OurDataset
  comprising more than 170k synthetic images of human bodies with various shapes and viewpoints
  as well as real (annotated) data for evaluation;
  (iii) we show that our model trained on synthetic data from scratch can perform well on real data,
  obtaining similar results to larger models initialized with pre-trained networks.
  It thus provides a good trade-off between performance and computation.
  Experiments on real data demonstrate the validity of our approach.
\end{abstract}

\section{Introduction}
\label{sec:intro}

Person detection and pose estimation are core components for multi-party Human-Robot Interaction (HRI).
%Particularly social robotics aims to provide the robot with social intelligence to autonomously interact with people and respond appropriately.
In particular, social robotics aims to provide the robot with social intelligence to autonomously interact with people and respond appropriately.
Detecting people in its surroundings and estimating their pose provide the robot the  means for fine-level motion understanding,
activity and behavior recognition, and in combination with other modalities, social scene understanding.
Although pose estimation has been widely studied, deploying fast and reliable systems remains a challenging task. On one hand, scenario's dynamic nature, \ie background clutter, multiple pose configurations, between people interaction, and sensing conditions may provoke partial observations hindering the detection process.
On the other hand, complex and accurate systems bring high computation burden, disabling the possibility for real-time deployment under limited computational budget.

\mypartitle{State-of-the-art.}
The classical method for body pose estimation is to model spatial relationships of body parts in a graphical model structure, provided part-specific detectors that perform over handcrafted features \cite{PoseMachines, GaussianProcessPose, PictorialStructures, PictorialRevisited}.
%,3DPictorialStructures}. 
Lately, Convolutional Neural Networks (CNN) have been proved to be an effective tool for pose estimation on RGB images. By means of a deep CNN, such a system detects human body parts in the image, which are subsequently parsed to produce body pose estimates.

\begin{figure}
	\includegraphics[width=.93\linewidth]{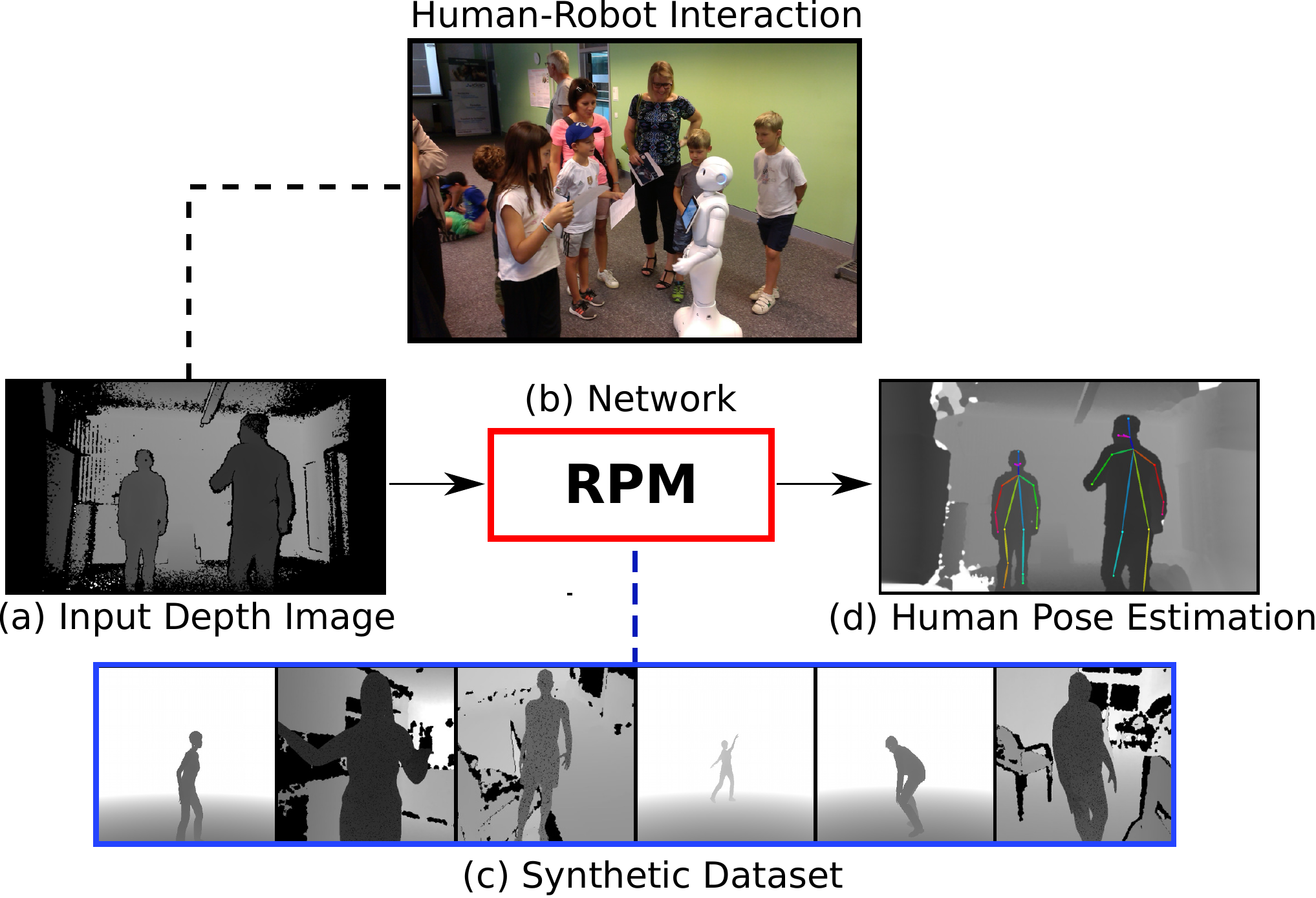}
	\vspace{-0.1cm}
	\caption{Overall scheme of the proposed method for efficient human pose estimation~(d) from depth images~(a).
          Our proposed \RPM convolutional network~(b) is trained with depth images consisting of synthetically generated people under multiple poses and positions
          combined with varying real background depth images~(c).
        }
\label{fig:scenario}
\vspace{-0.15cm}
\end{figure}

%A typical way to address pose estimation with CNN is inspired by the cascade of detectors concept, \ie sequentially stacked detectors (blocks of convolutional layers) to improve and refine body part predictions. Spatial image context is retrieved by various kernel resolutions \cite{CPM, DeeperCut, DeepCut, DepthMultitask} or embedding coarse to fine prediction in the network architecture \cite{HourGlass, BottomUpTopDown, CNNPartHeatmap}.
%
%Pose inference becomes hard for multi-person scenarios, since the number of possible pose configurations grows with the number of body part hypothesis.

% MV: some changes.
A conventional way to address pose estimation with CNN is inspired by the
cascade of detectors concept. That is, sequentially stacking detectors (blocks
of convolutional layers) to improve and refine body part predictions using
spatial image context. Image context is retrieved by various kernel
resolutions \cite{CPM, DeeperCut, DeepCut, DepthMultitask} or embedding coarse
to fine prediction in the network architecture \cite{HourGlass,
BottomUpTopDown, CNNPartHeatmap}.

% MV: I think this does not connect with the story.
%Pose inference becomes hard for multi-person scenarios, since the number of possible pose configurations grows with the number of body part hypothesis.

%%%%%%%%%%%%%%%%%%%%%%%%%%%%%%%%%%%%%%%%%%%%%%%%%%%%%%%%%%%%%%%%%%%%%%
% Figure of image synthesis placed here to be on the second page
%%%%%%%%%%%%%%%%%%%%%%%%%%%%%%%%%%%%%%%%%%%%%%%%%%%%%%%%%%%%%%%%%%%%%%
%\begin{figure*}[h!]
%\centerline{
%\hspace*{-2mm}
%a)\includegraphics[width=7.4cm]{./imgs/characters2.png}
%b)\hspace*{-3mm}\includegraphics[width=1.3cm]{./imgs/skeletonModel.png}
%c)\hspace*{-0mm}\includegraphics[width=3.5cm]{./imgs/sampleMummer3.jpg}
%d)\includegraphics[width=4.4cm]{./imgs/bg-fuse.png}
%}

%\caption{(a) Sample 3D characters with different poses and outfits;
  %%(b) Synthesizer scenario configuration. Three cameras are randomly placed and towards the 3D
  %(b) skeleton model;
  %(c) rendered synthetic depth image sample;
  %(d) examples  of training images, combining synthetic generated bodies with real background images.
  %}
%\label{fig:randomscenario}
%\end{figure*}

\begin{figure*}[h!]
	\center
	\includegraphics[width=.98\linewidth]{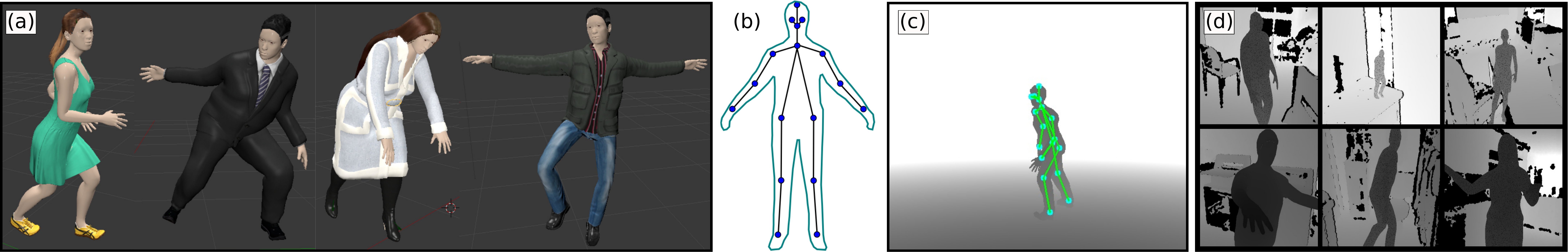}
	\vspace{-0.1cm}
	\caption{(a) Sample 3D characters with different poses and outfits;
	%(b) Synthesizer scenario configuration. Three cameras are randomly placed and towards the 3D
	(b) skeleton model;
	(c) rendered synthetic depth image sample;
	(d) examples  of training images, combining synthetic generated bodies with real background images.
	}
\label{fig:randomscenario}
\end{figure*}

%%%%%%%%%%%%%%%%%%%%%%%%%%%%%%%%%%%%%%%%%%%%%%%%%%%%%%%%%%%%%%%%%%%%%%

Spatial relationships between pairs of body parts are also considered in order to improve estimation and ease the inference stage.
%The relationships can be modeled by explicit regressors \cite{ArtTrack, PoseTrack}, or embedded in a neural network architecture \cite{TompsonLecun, CPMPaf}.
These relationships can be modeled by explicit regressors \cite{ArtTrack, PoseTrack}, or embedded in a network architecture \cite{TompsonLecun, CPMPaf}.
%Inspired by the cascade of detectors concept, \cite{CPMPaf} stacks detector blocks to refine predictions and encodes  body parts pairwise dependencies as a vector field between adjacent parts.
Motivated by the cascade of detectors concept, \cite{CPMPaf} relies on recurrent detector blocks to refine predictions and encode body parts pairwise dependencies as a vector field between adjacent parts.
Body landmarks\footnote{In this paper we use body parts and landmarks interchangeably.} and pair-relationships are learned in an end-to-end fashion and jointly predicted in a multi-task approach.
%Although research has succeeded in building very deep CNN models and achieving excellent results, the computation capabilities demands grow with the network's depth. In addition, large amounts of data for learning are required in order to prevent the model from overfiting.
Although very deep network models like \cite{CPMPaf} have provided excellent results, the computational demands of these models
grow with the network's depth and require large amounts of training data to prevent overfiting.

Depth data has also been used for pose estimation \cite{ShottonACM, DepthGestureSVM, RGBDDetection, MultiView3D, GaussianKernel3D}.
%As depth image encodes the distance between the camera and the objects in the scene,
Indeed, depth discontinuities and variations preserve many essential features that also appear in natural images like corners, edges, silhouettes.
It is also texture and color invariant, which may help to remove ambiguities in scale and silhouette shapes.
%Several approaches that use hand crafted features in combination with classical machine learning methods have been proposed
%The most well-known approach for pose estimation from depth images was presented by Shotton et al \cite{ShottonPAMI, ShottonACM}.
%Using simple depth-based handcrafted features, random forest are learned to pixel-wise label the image as belonging to one of the different body parts.
In \cite{ShottonACM}, for example, a random forest based on simple depth features is computed to pixel-wise label the image as belonging to one of the different body parts.
The need for training data was addressed by synthesizing depth images  with a large variety of human shapes and poses using computer graphics.
%However, the method assumed background subtraction as a preprocessing step, was limited to near frontal pose and close range observations.
Despite the remarkable and real-time results, the method assumes background subtraction as a preprocessing step, and is limited  to near-frontal pose and close-range observations.

%CNN-based methods have also been proposed for articulated pose estimation from depth images.
%Different approaches aim to regress body part likelihood on the image \cite{SyntheticHumans, 3DSighgraph, Deep3DPose} or to classify patches as body parts provided handcrafted features \cite{DepthInvariant, DepthMultitask, SkeletonFree, Hand3DPose}.
%These methods normally use an already pre-trained and large network architecture to train and deploy on pose estimation.
%Thus, they keep the network structure that normally has many parameters, not always appropriate for the problem, and unnecessary increases the time of the network's forward pass.

CNN-based methods have also been proposed for articulated pose estimation
from depth images~\cite{DepthInvariant, DepthMultitask, SkeletonFree}.
However, these methods normally use an already pre-trained and large network
(\eg VGG~\cite{VGG_Simonyan_2014}) as feature extractor to perform, subsequently,  the
prediction of human body landmarks.  As a consequence, they are not
appropriate for real-time pose estimation since such  pre-trained networks
involves many parameters and increases the computational cost during runtime.

\mypartitle{Approach and contributions.}
%
%Inspired by current advances in CNN, we investigate NN architectures that perform on depth images for fast and
Inspired by current advances in CNN, we investigate network architectures that perform on depth images for efficient and
reliable pose estimation in social multi-party HRI applications, as illustrated in Figure~\ref{fig:scenario}.
%We present a randomized synthesis pipeline built upon computer graphics software in order to produce synthetic depth images for training purposes. We design a small network architecture that predicts confidence maps directly on the depth image, modeling body part pairwise relationships, at low computational cost.
%
Depth data provides direct and very relevant information for body landmark detection like head, shoulders, or arms,
although the lack of texture may limit its performance where only subtle depth variations are expected (eyes, arms on body).
Also, thanks to the depth, moving from landmark localization to the actual 3D body pose will be more straightforward than with
RGB images only.
The challenge addressed in this paper is thus to gain speed  without sacrifying performance. In that direction, our contributions are:
\begin{compactitem}
\item we propose a fast and efficient network based on residual blocks,
  called Residual Pose Machines (\RPM), for body landmark localization from depth images;
  %, \MV{see Figure~\ref{fig:scenario}(b)};
%
\item we built a dataset of Depth Images of Humans (\OurDataset) comprising more than 170K synthetically generated depth images of humans,
  and which can be used for training purposes, along with 460 depth real images annotated with body landmarks.
  The dataset will be made publicly available;
\item we demonstrate that models trained on synthetic data can perform well on real data.
%
%\item we show that shallow models trained from scratch obtain similar results as larger models initialized with pretrained architectures,
\item we show that our relatively shallow \RPM model trained from scratch obtain similar results to larger models initialized with pre-trained networks,
thus providing a good trade-off between performance and computation.
\end{compactitem}

% \MV{* Above.. the title is approach and contributions... but the approach is not explained}

% \MV{* I think this contribution of "investigating network architectures" should be changed to "we propose an efficient network based on residual block, called RPM", or something like that..}

%In Section~\ref{sec:synthesis} we describe our synthesis pipeline to build synthetic depth images for training purposes. Section~\ref{sec:netdetails} decribes the CNN architecture design we chose for fast pose estimation. The experiments and obtained results are described in Section~\ref{sec:results}. In Section~\ref{sec:conclusions} we discuss conclusions and future work.

In Section~\ref{sec:synthesis}, we present our pipeline to build synthetic
depth images for training the network. Section~\ref{sec:netdetails} describes
the proposed network for efficient pose estimation. Experiments and results are
described in Section~\ref{sec:results}. Finally, Section~\ref{sec:conclusions}
concludes the paper.

\section{Synthetic Depth Image Generation}
\label{sec:synthesis}

%%%%%%%%%%%%%%%%%%%%%%%%%%%%%%%%%%%%%%%%%%%%%%%%%%%%%%%%%%%%%%%%%%%%%%
% Image of architecture placed here to be on page 3
%%%%%%%%%%%%%%%%%%%%%%%%%%%%%%%%%%%%%%%%%%%%%%%%%%%%%%%%%%%%%%%%%%%%%%
\begin{figure*}
\centering
\includegraphics[width=1\linewidth]{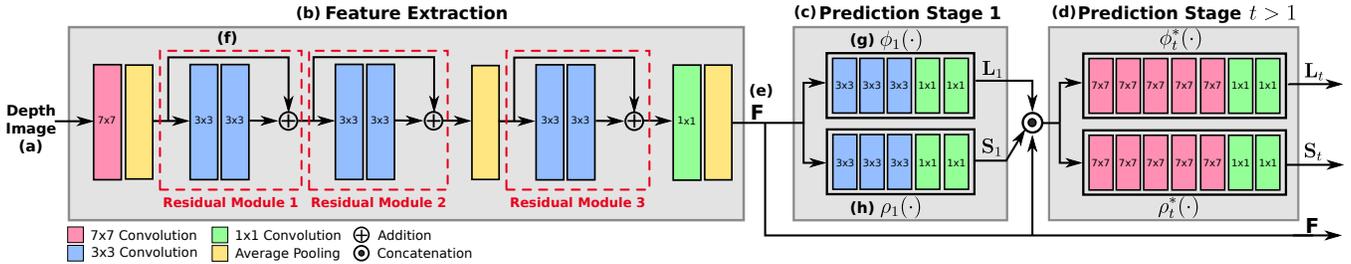}
\caption{Architectural design for the Residual Pose Machines
  (\RPM). The input to the network is a single channel
	depth image~(a). The feature extractor~(b) is composed of three residual
	modules~(f) producing \NFeatureChannels feature channels. The
  branches $\phi_t$ and $\rho_t$~(c,d) predict confidence maps for the
  location of body parts and limbs.
}
\label{fig:res-arch}
\end{figure*}
%%%%%%%%%%%%%%%%%%%%%%%%%%%%%%%%%%%%%%%%%%%%%%%%%%%%%%%%%%%%%%%%%%%%%%

% %%%%%%%%%%%%%%%%%%%%%%%%%%%%%%%%%%%%%%%%%%%%%%%%%%%%%%%%%%%%%%%%%%%%%%
% \begin{figure*}[!h]
%   \centering
%   \hspace*{-2mm}
% \subfloat[]{
% \includegraphics[width=7.5cm]{./imgs/characters2.png}
% }
% \subfloat[]{
% 	\includegraphics[width=1.4cm]{./imgs/skeletonModel.png}
% }
% \subfloat[]{
% 	\includegraphics[width=3.6cm]{./imgs/sampleMummer3.jpg}
% }
% \subfloat[]{
% 	\includegraphics[width=4.6cm]{./imgs/bg-fuse.png}
% }

% \caption{(a) Sample 3D characters with different poses;
%   %(b) Synthesizer scenario configuration. Three cameras are randomly placed and towards the 3D
%   (b) skeleton model;
%   (c) rendered synthetic depth image sample;
%   (d) examples  of training images, combining synthetic generated bodies with real background images.
%   }
% \label{fig:randomscenario}
% \end{figure*}
% %%%%%%%%%%%%%%%%%%%%%%%%%%%%%%%%%%%%%%%%%%%%%%%%%%%%%%%%%%%%%%%%%%%%%%

% We inspired in \cite{ShottonPAMI} and implement a randomized synthesizer with our target scenario configuration.

Training CNNs requires large amounts of data with annotations.
Unfortunately, a precise manual annotation of depth images with body parts is not so easy, given that people roughly appear as blobs.
Fortunately, as shown by Shotton \etal \cite{ShottonACM}, synthesizing depth images of the human body is easier than synthesing real RGB images,
since color, texture, and illuminations conditions are much more difficult to render in practice.
%Computer graphics software provides means by which the depth in a rendering scene can be extracted and easily produce realistic depth images.
%Therefore, we chose to address the lack of training by synthesizing depth images using computer graphics.
In this paper, we follow this approach.
Rougly speaking, we follow a randomized synthesis pipeline: we created a dataset of 3D human characters,  took real motion capture (mocap) data to re-target
their pose in the 3D space, selected several random viewpoints, added real backgrounds, and generated the ground-truth, 
as illustrated in Figure~\ref{fig:randomscenario}.
The main issues are how to produce images with enough variations in human shapes, body pose and viewpoint configurations,
how to (automatically) annotate these images, and how to simulate realistic backgrounds.
This is detailed below.

%% Depth image synthesis brings the following benefits: high quality annotations are automatically extracted at rendering time, and variation in human shapes, body pose configurations, and viewpoints in the target scenario are easily introduced. This facilitates the acquisition of data specific to our human-robot interaction scenario, \ie people mainly in upright-frontal pose. To synthesize realistic depth images we need to cover two potential problems:
%% \begin{enumerate}
%% 	\item human body exhibit a large range of configurations; synthetic data has to span this pose space, and
%% 	\item physical features and clothing increase the variability in the observed body shapes.
%% \end{enumerate}
%% To account for variation in body shape and body configurations, we  We describe our synthesis pipeline in the next sections.

\mypartitle{Variability in body shapes}.
We built a dataset of 24 adult 3D characters using the modeling software Makehuman \cite{Makehuman}.
Characters are of both genders and with different heights and weights,
and  have been dressed with different clothing outfits to increase shape variation
(skirts, coats, pullovers, \etc), see Figure~\ref{fig:randomscenario}(a) for a better illustration.

\mypartitle{Variability in body poses}.
For each character, we performed motion retargeting from motion capture data.
We relied on the publicly available motion capture database of CMU labs~\cite{CMUMocap},
selecting motion capture sequences grossly fitting our scenario in which the robot interacts with people appearing mainly in an upright configuration:
people standing still, walking, turning, bending, picking up objects, \etc.

\mypartitle{Variability in view point.}
We placed the 3D character in a reference point with a reference orientation, and defined a recording zone as a circle of 8m radius centered at the character.
Then, a camera was randomly placed (position and height) in this zone, and its orientation was defined by randomly selecting a point on the character torso and
pointing the camera to it.
%
%We split the zone into three subzones and randomly place a camera in each \ie accounting for view position. The orientation of each of the cameras is selected by randomly selecting a point on the character torso and pointing the camera to it. The orientation and position of the camera is kept fixed during rendering, whereas we add random perturbations to the character orientation to increase the variability of the viewpoints. Fig.~\ref{fig:randomscenario}(b) shows our synthesizer configuration.
Pose retargeting and depth buffer rendering were performed using the computer graphics software Blender \cite{Blender},
observe Figure~\ref{fig:randomscenario}(c) for an example.

\mypartitle{Dataset and annotations.}
The synthetic images of the \OurDataset public dataset were generated with our synthesis 
pipeline~\footnote{{\color{red}\url{https://www.idiap.ch/dataset/dih}}}.
They comprise 172148 images of a single person performing different types of motion under different viewpoints. 
As the human characters come along with 3D semantic joint locations, we were able to automatically
record the location  of  17 body landmarks (\emph{head, neck, shoulders, elbows, wrists, hips, knees, ankles, eyes}, see Figure~\ref{fig:randomscenario}(b),
in the world, camera, and image coordinate systems using the camera's calibration and projection matrices.
Each of the 17 keypoints was then automatically labeled as visible or invisible by thresholding the distance between
the keypoint and the body surface point closest to the camera located on the line between the keypoint and the camera.
In addition, the silhouette's mask was also extracted to allow the incorporation of the body depth images
in real depth images (cf below and Section~\ref{sec:expdata}).
%
%The dataset will be made publicly available for research purposes.

\mypartitle{Adding real background.}
Note that a realistic dataset needs as well realistic background content.
Rather than generating a predefined set of images with random (real) background, such images
were produced on the fly during training, as described in Section~\ref{sec:expdata}.
Some example images are shown in Figure~\ref{fig:randomscenario}(d).

\section{Depth-based Pose Estimation Approach}
\label{sec:netdetails}

% %%%%%%%%%%%%%%%%%%%%%%%%%%%%%%%%%%%%%%%%%%%%%%%%%%%%%%%%%%%%%%%%%%%%%%
% \begin{figure*}
% \centering
% \includegraphics[width=1\linewidth]{./ink_drawings/arch_new}
% \caption{Architectural design for the Residual Pose Machines
%   (\RPM). The input to the pose regressor network is a single channel
%   depth image. The feature extractor is composed by three residual
%   modules that produce \NFeatureChannels feature channels. The
%   branches $\phi_t$ and $\rho_t$ predict confidence maps for the
%   location of body parts and limbs.}
% \label{fig:res-arch}
% \end{figure*}
% %%%%%%%%%%%%%%%%%%%%%%%%%%%%%%%%%%%%%%%%%%%%%%%%%%%%%%%%%%%%%%%%%%%%%%

Our model is inspired by  the Convolutional Pose Machines~(CPM) \cite{CPMPaf} approach,
which builds a powerful CNN-based 2D body pose detector for color images
trained to jointly localize the body parts and limbs of multiple people.
%The efficient part of the CPM is that it leverages the context by iteratively
%refining its own predictions.
%
% The CPM is a CNN which consists of a feature extractor
% part (VGG network), followed by two branches, one predicting the
% location of body parts (nose, eyes, ankle, \etc), and one predicting
% the limbs (forearms, forelegs, \etc). Each branch is made of several
% stages which iteratively refines the predictions of the previous
% stage. Pose inference is then performed in a greedy bottom-up step to
% gather parts and limbs belonging to the same person.
%
% parsing body part observations in a sparse connected tree in a
% pairwise manner. The power of this method lies on its multi-task
% approach for predicting body parts and body limbs. By explicitly
% modeling the body part relationships through the vector fields, the
% network learns the possible configurations in which two body parts are
% connected. In addition, the network learns to exploit the spatial
% context between body parts and limbs by stacking blocks of
% convolutional layers to refine the predictions from previous blocks.
%
In this section we present our network model, %how we adapt the CPM architecture
with the aim of reducing the number of parameters and speeding up the whole process for robotics applications.

\subsection{Architecture}

% \subsection{Prediction refinement by detector blocks}
\mypartitle{Overview.}
Figure~\ref{fig:res-arch} depicts the architecture of the proposed efficient
network, dubbed Residual Pose Machines (\RPM), to detect body parts and limbs, 
%This architecture is adapted from~\cite{CPMPaf},
and which takes as input a single channel depth image.

More precisely, the input depth image, see Figure~\ref{fig:res-arch}(a), is fed
into a feature extraction module to get a compact and discriminative feature
representation denoted as~$\mathbf{F}$ (refer to
Figure~\ref{fig:res-arch}(b,e)).
Then, these features are passed to a series of prediction stages (Figure
~\ref{fig:res-arch}(c,d)) in order to localize in the image the body landmarks
(nose, eyes, ankle, \etc) and limbs (segments between two landmarks according
to the skeleton shown in Figure~\ref{fig:randomscenario} such as forearms,
forelegs, \etc).

Each prediction stage consists of two branches made of fully convolutional
layers. The first branch, denoted as~$\rho_t(\cdot)$, is trained to localize
the body parts (Figure~\ref{fig:res-arch}(h)), while the second
one~$\phi_t(\cdot)$ is trained to localized the body limbs
(Figure~\ref{fig:res-arch}(g)). The prediction stages are applied
sequentially with the goal of refining the predictions of body parts and limbs
using the result of the previous stage and incorporating spatial image context.

%The input depth image, see Figure~\ref{fig:res-arch}(a), is processed by a CNN
%to produce a set of features denoted as~$\mathbf{F}$
%The features are then processed to localize the
%body landmarks (nose, eyes, ankle, \etc) and the body limbs (segments between 2
%landmarks according to the skeleton in Figure~\ref{fig:randomscenario}, like
%forearms, forelegs, \etc).  They pass in a succession of stages, each stage
%refining the result of the previous one.  A stage consists of two branches made
%of one fully convolutional network. The first branch, denoted
%as~$\rho_t(\cdot)$, is trained to localize the body parts, while the second
%branch~$\phi_t(\cdot)$ is trained to localized the body limbs.

Finally, pose inference for \RPM is performed in a greedy bottom-up step to
gather parts and limbs belonging to the same person, in the same way as for
CPM. Figure~\ref{fig:output} shows some results of \RPM on depth images.

\mypartitle{Feature extraction network.} \label{subsec:shallow}
Depth images exhibit less details than color images (\ie color and texture),
and posture information mainly lies in the person silhouette in combination
with the body depth surface.
%, although parts such as the eyes are more difficult to detect with less
%texture.
This motivates for the use of a smaller network architecture compared to those
used for color images.
%like the VGG network.
Therefore, rather than relying on the VGG-19 network used in~\cite{CPMPaf} as
feature extractor, we propose to use the smaller and lightweight network
architecture shown in Figure~\ref{fig:res-arch}(b).
%We adapt the feature extractor (originally a VGG-19) with as feature extractor
%the smaller network shown in Figure~\ref{fig:res-arch}.
%
It consists of an initial convolutional layer followed by three residual
modules (or blocks)~\cite{he2016deep} with small kernel sizes of ${3 \times 3}$.
The network has three average pooling layers.
Each residual module, see Figure~\ref{fig:res-arch}(f), consists of two
convolutional layers and a shortcut connection (hence the name 'residual'~\cite{he2016deep}:
the inner part of the module is supposed to only model the incremental information since the shortcut
represents the identity mapping). Batch normalization and ReLU
are included after each convolutional layer and after the shortcut connection.

Our motivation to use residual blocks is that they are known to outperform VGG
networks, and to be faster by having a lower computational
cost~\cite{canziani2016analysis}.

\mypartitle{Confidence maps and part affinity fields prediction.}
\label{subsec:paf}
The feature extractor is followed by a succession of stages, each stage taking
as input the features $\mathbf{F}$ and the output of the previous stage. As
depicted on Figure~\ref{fig:res-arch} and mentioned before, a stage consists of
two branches of convolutional layers, the first branch predicting the location
of the parts, and the second predicting the orientation of the limbs. We keep
the same design of the branches $\phi_t(\cdot)$ and $\rho_t(\cdot)$ as in the
original CPM~\cite{CPMPaf} to maintain the effective receptive field as large
as possible. That is, in the first prediction stage the network has three
convolutional layers with filters of ${3 \times 3}$ and two layers with filters
of ${1 \times 1}$, whereas in the remaining stages there are five and two
convolutional layers with filters of ${7 \times 7}$ and ${1 \times 1}$,
respectively.

%the network are made of three filters
%of ${3 \times 3}$ and two filters of ${1 \times 1}$ convolutional layers, and
%in the remaining stages, we have five ${7 \times 7}$ and two ${1 \times 1}$
%convolutional layers.

Table~\ref{table:cpmvsres} shows a comparison of the number of parameters for
different designs of \RPM and CPM. Note that with only one stage there is no
refinement since the first stage only takes as input the features.
%Specifically Table~\ref{table:cpmvsres}, RPM-1S (1 stage) does not refine,
Specifically, RPM-1S denotes RPM with one stage while RPM-2S corresponds to the
network with two prediction stages (refinement).

\subsection{Training and confidence map ground truthing}

We regress confidence maps for the location of the different body
parts and predict vector fields for the location and orientation of
the body limbs.
In this section, for simplicity, we  follow the original notation of~\cite{CPMPaf}.
The ideal representation of the body part confidence
maps $\mathbf S^{*}$ encodes the locations on the depth image as
Gaussian peaks. Let $\mathbf x_j$ to be the ground truth position of
body part $j$ on the image. The value for pixel $\mathbf p$ in the
confidence map is computed as follows
\begin{equation}
\mathbf S^{*}_j(\mathbf p) = exp \left(-\frac{|| \mathbf p_j - \mathbf x_j ||_2^2}{\sigma}  \right),
\label{eq:peaks}
\end{equation}
where $\sigma$ is empirically chosen.

% In the case that there are $K$ people in the image, the confidence
% map to be predicted is the aggregation of confidence maps across $K$
% people $S_j^{*}(\mathbf p) = \max_k S_{jk}^{*}(\mathbf p)$.

The ideal representation of the limbs $\mathbf L^{*}$ encodes the
confidence that two body parts are associated, in addition to
information about the orientation of the limbs by means of a vector
field. Consider a limb of type $c$ that connects two body parts $j_1$
and $j_2$, \eg elbow and wrist, with positions on the depth image
$\mathbf x_{j_1}$ and $\mathbf x_{j_2}$. The ideal limb affinity field
at point $\mathbf p$ is defined as
\begin{equation}
\mathbf L_c^{*}(\mathbf p) =
	\begin{cases}
	\mathbf v, & \text{if $\mathbf p$ on limb $c$},\\
	0         &  \text{otherwise},
	\end{cases}
\end{equation}
where $\mathbf v$ is the unit vector that goes from $\mathbf x_{j_1}$
to $\mathbf x_{j_2}$. The set of pixels that lie on the limb are those
within a distance to the line segment that joins the two body parts.

% For $K$ people in the image, the vector field to be predicted is the
% average of non zero vectors across all people
% ($\mathbf L_c^{*}(\mathbf p) = \frac{1}{n_p} \sum_k \mathbf
% L_{ck}^{*}(\mathbf p)$).

%Intermediate supervision is applied at the end of each block to
%prevent from vanishing gradients. This supervision is implemented by
Intermediate supervision is applied at the end of each prediction stage to
prevent the network from vanishing gradients. This supervision is implemented by
two $L_2$ loss functions, one for each of the two branches, between
the predictions $\mathbf S_t$ and $\mathbf L_t$ and the ideal
representations $\mathbf S^{*}$ and $\mathbf L^{*}$. The loss
functions at stage $t$ are
\begin{eqnarray}
f_t^1 = \sum_{\mathbf p \in \mathbf I} || \mathbf S_t(\mathbf p) - \mathbf S^{*}(\mathbf p)  ||_2 ^2, \\
f_t^2 = \sum_{\mathbf p \in \mathbf I} || \mathbf L_t(\mathbf p) - \mathbf L^{*}(\mathbf p) ||_2 ^2.
\label{eq:losses}
\end{eqnarray}
The final multi-task loss is computed as
$f = \sum_{t=1}^T \left( f_t^1 + f_t^2 \right)$ where $T$ is the total
%number of iteration blocks.
number of network stages.

\subsection{Implementation details}

\mypartitle{Image preprocessing.}
The depth images are normalized by scaling linearly the depth values
in the $[0,8meter]$ range into the  $[-0.5,0.5]$ range.
%dividing the depth values by 8~meters.
% We encode the depth values by normalizing the depth image in the
% range $[-0.5,0.5]$ by dividing the depth values by 8m and shift it
% accordingly.
%
Furthermore, note that the real data contains noise and missing
values, especially around body silhouette due to the sensing process
(see Figure~\ref{fig:res-arch}). Although more advanced domain
adaptation techniques could be used to reduce the mistmatch between
the clean synthetic data and the noisy real data distributions
\cite{Ganin_DANN, GANApple}, in this paper we considered using a
%simple inpainting preprocessing (from the OpenCV library) to fill out
simple inpainting preprocessing to fill out
the noise and shadows around the body silhouette and thus prevent
sharp discontinuities to potentially affect the network output.
This is shown in the experimental section.

%\JM{Add citation above}
\mypartitle{Network training.}  Pytorch is used in all our experiments.  We
train different network architectures with stochastic gradient descent with
momentum for 100K iterations each.  We set the momentum to 0.9, the weight
decay constant to ${5 \times 10^{-4}}$, and the batch size to~10.
%for all experiments.
We uniformly sample values in the range $[4 \times 10^{-10}, 4 \times 10^{-5}]$
as starting learning rate and decrease it by a factor of 10 when the training
loss has settled.
All networks are trained from scratch and progressively, \ie to train
%network architectures with $t$ block detectors, we initialize the
network architectures with $t$ stages, we initialize the network with the
parameters of the trained network with ${t-1}$ block detectors.

\mypartitle{Part association.}
We use the algorithm presented in \cite{CPMPaf} that uses the part
affinity fields as confidence to associate the different body parts
and perform the pose inference.

% Results

\section{Experiments and Results}
\label{sec:results}

\begin{figure*}[!h]
\centering
\includegraphics[width=0.98\textwidth]{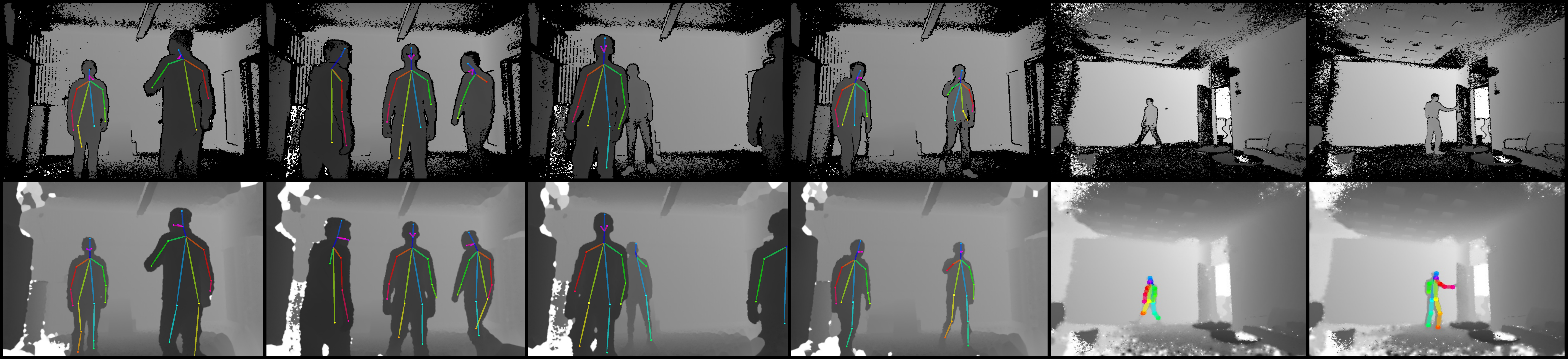}
\vspace{-0.1cm}
\caption{Output of the proposed RPM-2S for some sequence instances of the testing set.
  Top: results on raw depth images.
  Bottom: results on depth images with inpainting.
 }
\label{fig:output}
\end{figure*}

We conducted experiments using the synthetic images of the \OurDataset dataset as well as on the real depth images of the dataset (described below).
We present the experimental protocol in Section~\ref{sec:protocol}, focusing on both accuracy and computational aspects.
The analysis of the results is presented in Section~\ref{sec:expresults},
where we study the impact  on performance of different modeling elements like network architecture or  preprocessing.

%We use the \OurDataset dataset to train various versions of our shallow CNN framework for pose estimation. We evaluate the model's performance in real depth images and compare it with the performance obtained by a wider and deeper network.

\subsection{Data preparation}
\label{sec:expdata}

\mypartitle{Synthetic data.}
We split the synthetic images of our \OurDataset dataset into three folds with the following percentage and amount of images:
training (85\%, 146327), validation (5\%, 8606), and testing (10\%, 17215).
%and validation, with corresponding sizes of 85\%, 10\% and 5\% out of the total data. Table \ref{table:datasize} summarizes the number of images per fold.

\mypartitle{Training data: data augmentation with real background images.}
Relying only on clean depth images of the human body may hinder the generalization capacity
of a trained network due to the data mismatch with  real images.
Thus, to avoid our pose detector to overfit clean synthetic image details, we propose to add perturbation to the synthetic images, and in particular,
to add real background content which will provide the network with real sensor noise.

\myparit{Adding real background content.}
Obtaining real background depth images (which do not require ground-truth) is easier than generating synthetic body images.
As backgrounds, we  consider the dataset in \cite{KinectNoise} containing 1367 real depth images recorded with a Kinect~1 and exhibiting depth indoor clutter.
We divided it into  three folds (training, validation and test) % folds of 85\%, 5\% and 10\% respectively,
which were associated with the corresponding synthetic body data folds.
Then, during learning, training images were produced on the fly
by randomly selecting one depth image background and body synthetic images,
and compositing a depth image using the character silhouette mask.
Care was taken to avoid conflicting depth data (\ie the character depth value should be in front of the background),
and with the character's feet lying on a flat surface.
Sample results are shown in 
%The result of this process is a depth image exhibiting a clean human silhouette with noisy real depth indoor background.
Fig.~\ref{fig:randomscenario}(d).

\myparit{Pixel noise.}
During training we randomly select 20\% of the body silhouette's pixels and set their value to zero.

\myparit{Image rotation.}
%An image in a batch is rotated with a probability 0.1 by a randomly selected
Training images are rotated with a probability 0.1 by a randomly selected
angle in the range $[-30, 30]$ degrees.

\mypartitle{Test data.}
To evaluate the performance of our algorithm, we rely on typical HRI real data captured with a Kinect~2 and
exhibiting one or multiple people people passing in front of or interacting with our robot.
%
%The data meet our conditions for HRI scenario.
We manually annotated the body landmarks in 460 periodically sampled images, resulting in a dataset of 546 person instances.

\subsection{Evaluation protocol}
\label{sec:protocol}

\mypartitle{Accuracy metric.}
We use standard precision and recall measures derived from the Percentage of Correct Keypoints (PCKh) evaluation protocol  as performance metrics \cite{Yang_PAMI}.
%
%The evaluation is performed as follows.
%
More precisely, after the forward pass of the network, we extract all the landmark predictions $p$ whose confidence are above a threshold $\ConfidenceMapThreshold$, and 
run the part association algorithm to generate pose estimates from these predictions\footnote{Note that in this algorithm, landmark keypoints not associated with any estimates
are automatically discarded}.
Then, for each landmark type, and for each ground truth points $q$, we associate the closest prediction $p$ (if any) whose distance to $q$
is within a distance threshold $\PCKhDistanceThresh=\PCKhCoef \times \PCKhBBGtHeight$, where \PCKhBBGtHeight stands for the height of the
bounding box of the person (in the ground truth) to which $q$ belongs to. 
Such associated $p$ are then counted as true positives,
whereas the rest of the landmark predictions
%within and outside the threshold $d$
are counted as false positives.
%
%Body part proposals without associated pose estimate are discarded.
%
%The closest prediction $p$ for which the distance to the ground truth point $q$ is inside a PCKh threshold $d$ is considered as true positive.
Ground truth points $q$ with no associated prediction are counted as false negatives.
%associated with a r  inside threshold $d$.
%
The average recall and precision values can then be computed by averaging over the landmark types and then over the dataset.
Finally, the average recall and precision values used to report performance are computed by averaging the above recall and precision
over several  distance thresholds \PCKhDistanceThresh by varying \PCKhCoef in the range $[0.05,0.15]$.
%
%% We can then compute the precision and recall for the different types of landmark keypoints, and then
%% average them for a given image and over the dataset.
%% %(percentage of 
%% %
%% %We modify the threshold $d$ in the range of $[0.01h, 0.2h]$, where $h$ stands for the height of the bounding box ground truth.
%% We average the precision and recall values among the thresholds $[0.05h,0.15h]$ to finally generate the recall-precision curve and compare the different network performances.

\mypartitle{Computational performance.}
Model complexity is measured via the number of parameters it comprises,
and the number of frames per second (FPS) it can process when considering only the forward pass of the network.
This was measured using the median time to process 2K images at resolution $444 \times 368$ with an Nvidia card GeForce GTX 1050.
%To that end, we proceed as follows: we take the time stamps just right before and after computing the forward pass of the network,
% we compute their difference and finally take the median time in seconds of the resultant distribution and compute the FPS accordingly.

\subsection{Tested models}

\mypartitle{Proposed model.}
Our pose detection model is built as in Section~\ref{sec:netdetails}. We configure the network parameters \NStages=$2$,
to profit from spatial context and \NFeatureChannels$=64$ to balance the speed-accuracy trade-off.
We refer to this configuration as \TheRPM.
In experiments, we will evaluate the impact of different parameters like:
the number of stages \NStages in the cascade of detectors part of the network;
the number of \NFeatureChannels  feature channels in the residual blocks,
and the impact of inpainting preprocessing step.

\mypartitle{CPM Baseline.}
We consider the original architecture presented in \cite{CPMPaf}, trained as our model with the \OurDataset data.
%We consider the architectures upto 2 detector blocks.
As in the original work, the architecture parameters are initialized using the first 10 layers of the VGG-19 network.
To accomodate the need for the 3 channel (RGB) image input expected by VGG-19, the single depth channel is repeated three times.
%We refer to this network as CPM in our experiments.

\begin{table*}[t]
\centering
\begin{tabular}{c c c c c c c c c}

\Xhline{2\arrayrulewidth}

\multirow{3}{*}{Architecture} & \multirow{3}{*}{Stages} & \multirow{3}{*}{$w$} & \multirow{3}{*}{N. Parameters} & \multirow{3}{*}{FPS} & \multicolumn{4}{c}{Performance} \\

\cline{6-9}

& & & & & \multicolumn{2}{c}{Complete body} & \multicolumn{2}{c}{Upper body} \\

\cline{6-9}

& & & & & AP & AR & AP & AR\\

\Xhline{2\arrayrulewidth}

Cao et al \cite{CPMPaf}  
           & 6  & --  & 51.86 M &  3.6 & 69.89*          & 67.43*         & 78.75*          & 78.10* \\

\Xhline{2\arrayrulewidth}

CPM-1S              & 1  & --  & 8.38 M  & 18.6 & 65.52          & 51.67          & 68.74         & 66.85\\

CPM-2S              & 2  & --  & 17.07 M & 11.2 & 70.36          & \textbf{57.03} & 76.00         & 71.77\\

RPM-1S              & 1  & 64  & 0.51 M  & 56.7 & 64.63          & 44.34          & 73.62         & 57.65\\

\textbf{RPM-2S}     & 2  & 64  & 2.84 M  & 35.2 & 65.46          & 56.77          & 74.86         & 71.96 \\

RPM-3S     			& 3  & 64  & 5.17 M  & 20.8 & 63.81          & 56.34          & 71.05         & 69.95 \\

\Xhline{2\arrayrulewidth}

RPM-1S           & 1  & 128 & 1.83 M  & 22.5 & 45.30             & 41.59          & 51.84          & 55.65\\

RPM-2S           & 2  & 128 & 10.5 M  & 12.5 & \textbf{72.19}    & 56.11          & \textbf{84.10} & \textbf{72.91}\\

\Xhline{2\arrayrulewidth}

\end{tabular}
\caption{Comparison of the performance and architecture components for the different tested network architectures.
  Note that results from \cite{CPMPaf} (marked with *) are provided for indication, as
  they are  computed over RGB images on a disjoint set of body landmark types than the one we use for depth.
  See Section~\ref{sec:results} for details.
}
\label{table:cpmvsres}
\vspace{-0.2cm}
\end{table*}

\begin{figure*}[tb]
\vspace{-0.2cm}

\centering
a)\includegraphics[width=0.3\textwidth]{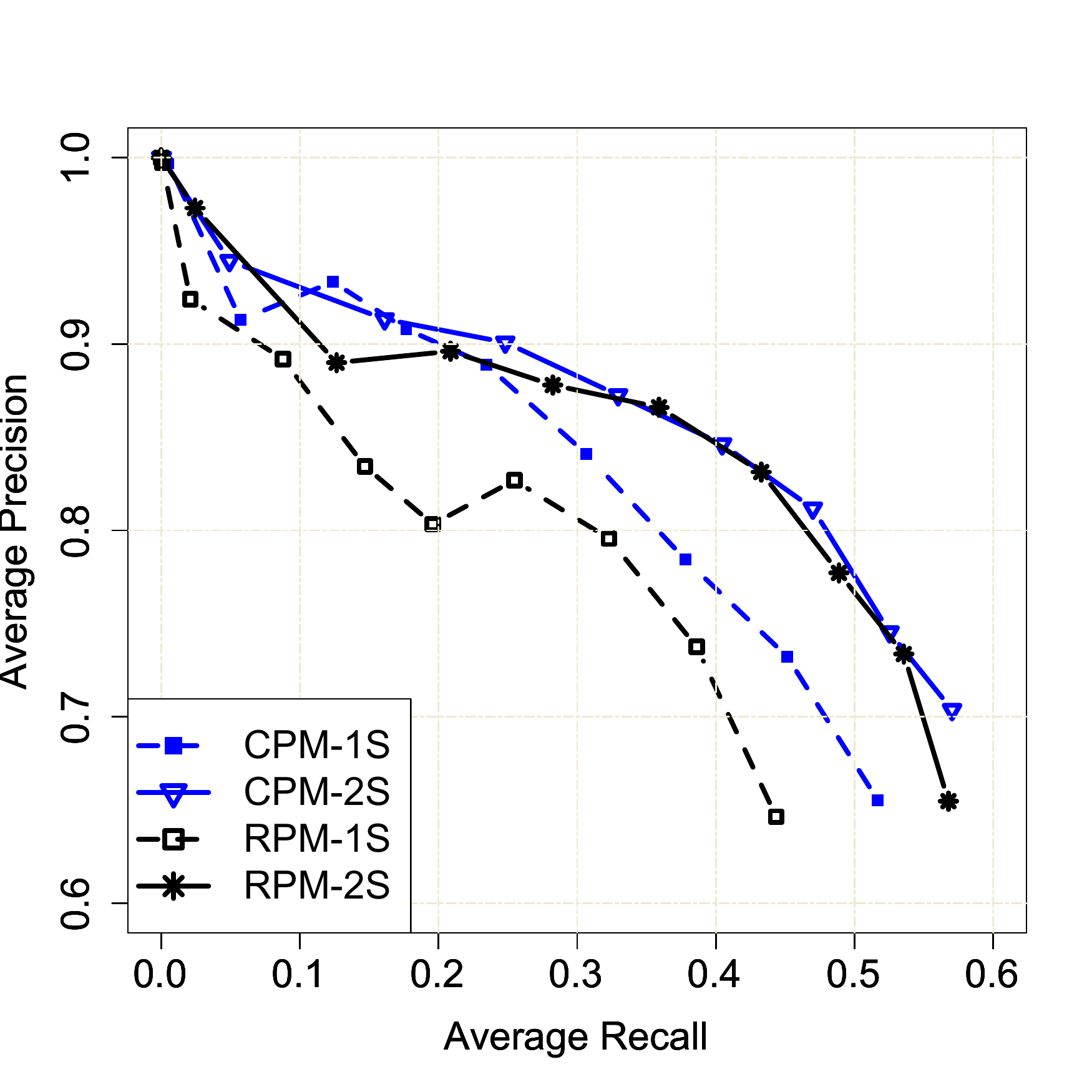}
b)\includegraphics[width=0.3\textwidth]{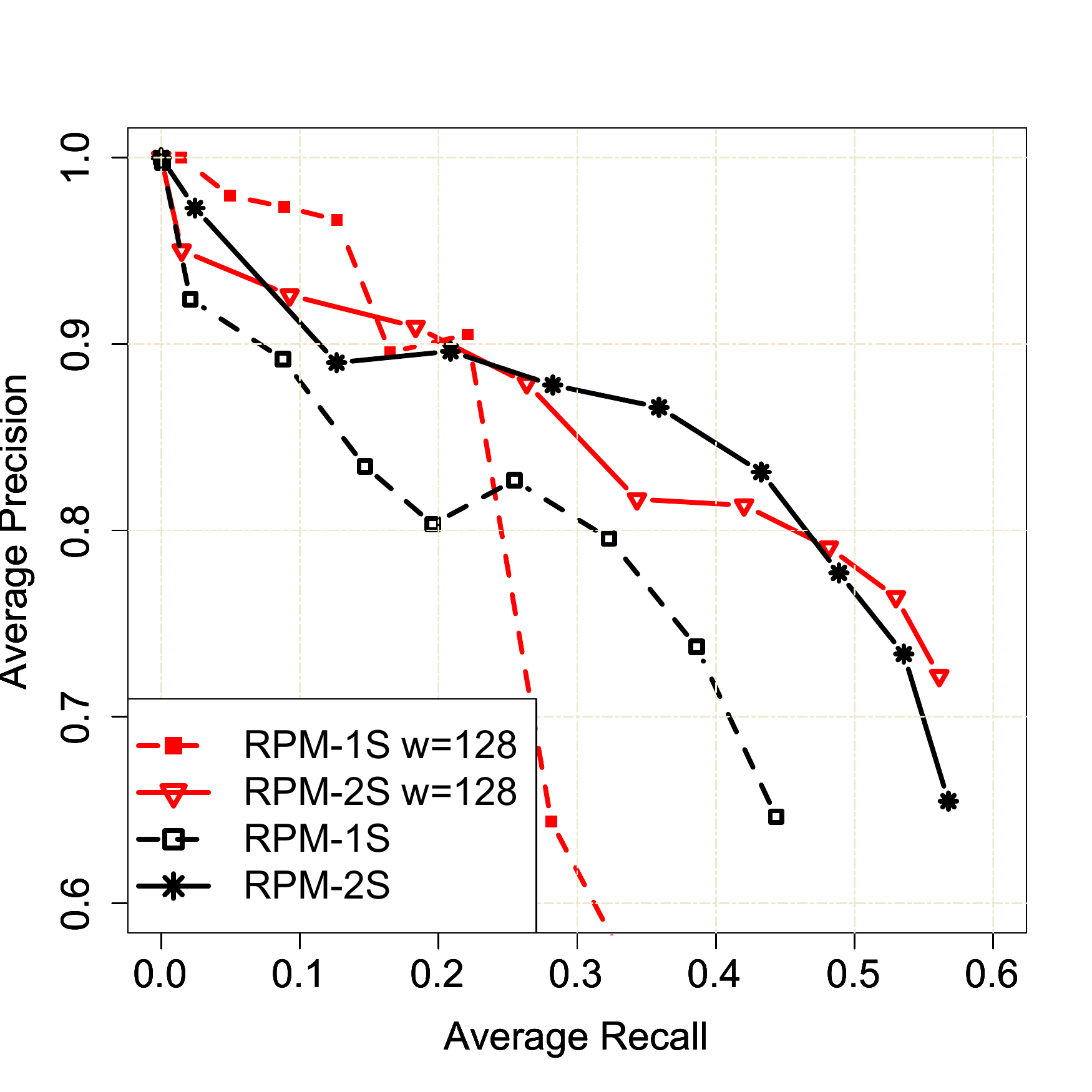}
c)\includegraphics[width=0.3\textwidth]{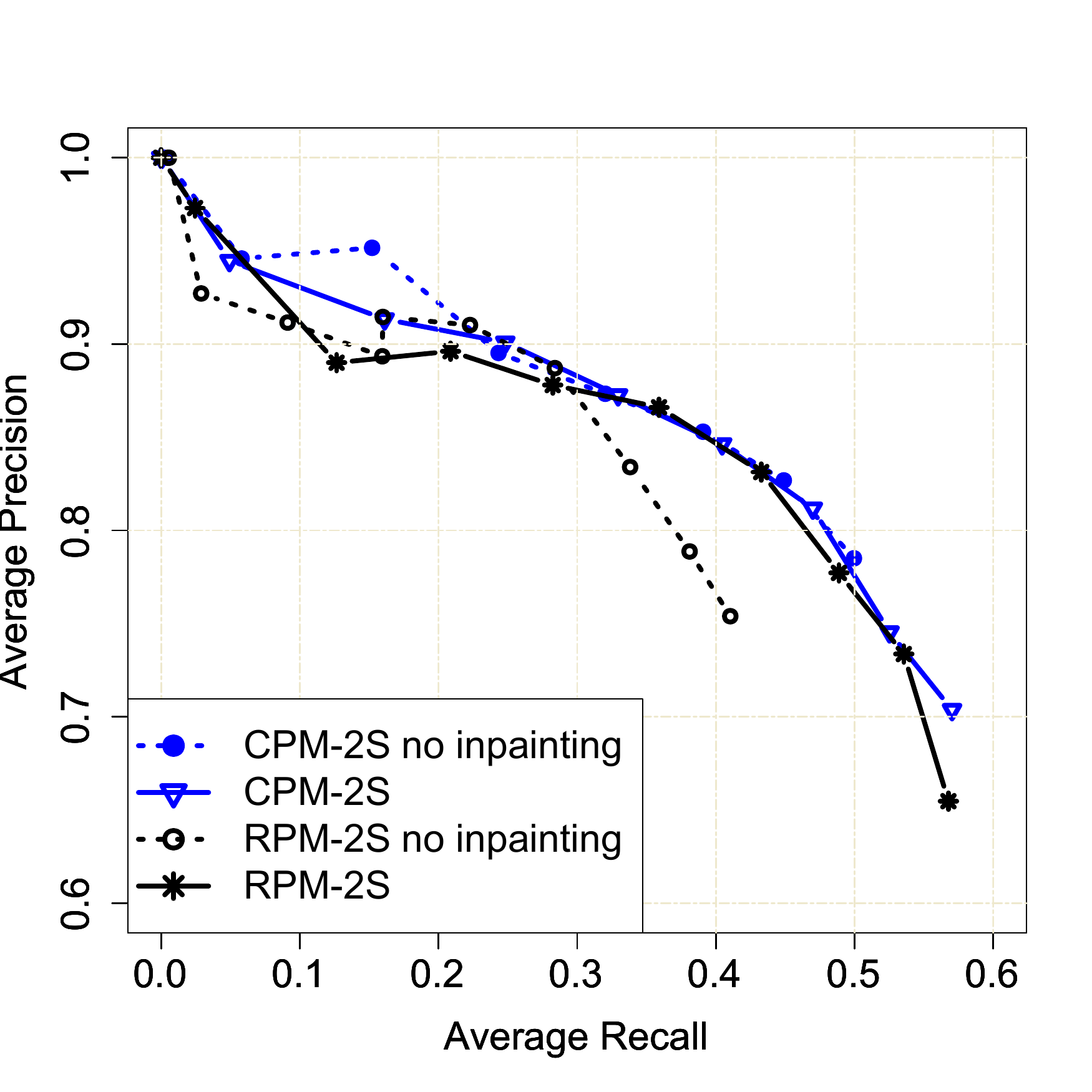}

\vspace{-0.2cm}
\caption{Average recall-precision curves.
  a) comparison between the baseline CPM and the proposed RPM networks, up to 2 prediction stages.
  b) impact of the number of feature channels \NFeatureChannels in the feature extractor CNN.
  c) impact of the  inpainting preprocessing.
}
\vspace{-0.2cm}
\label{fig:recall-prec-cmp}
\end{figure*}

\subsection{Results}
\label{sec:expresults}

Table \ref{table:cpmvsres} compares the performances of the different methods.
We report both the average recall and precision for all landmark types in the complete skeleton model
(see Fig.~\ref{fig:randomscenario}b)), and for the upper body, \ie \emph{head, neck, shoulders, elbows and wrists}, since upper-body detection might be sufficient for most
typical HRI application.
The table also compares the FPS and the number of trainable parameters of the different networks.
As an extra experiment, for comparison, we show the results obtained by running the code of  \cite{CPMPaf} over the registered RGB images.
For this experiment, the performances were computed over the body parts that the skeleton model in \cite{CPMPaf} has in common with our skeleton model.
In addition, Figure~\ref{fig:recall-prec-cmp} reports precision-recall curves obtained by varying the \ConfidenceMapThreshold threshold which impacts the number of detected
keypoints before the body part association step.

\mypartitle{Analysis.}
From this table, we can see that our  proposed network \TheRPM ($\NFeatureChannels=64$) performs as well as the baseline CPM-2S
(\eg for upper-body, it has a recall-precision of $72$ and $74.8$ vs $71.7$ and $76$ for CPM-2S)
but with 6 times less parameters and being 3.14 times faster.
Interestingly, we can notice from Fig.~\ref{fig:recall-prec-cmp}a that without using recursion (\RPM-1S), the smaller complexity of our feature extraction CNN indeed leads
to degraded performance compared to the original CPM-1S, but that this gap is filled once the recursion is introduced (compare the \RPM-2S and CPM-2S curves). 

\myparit{Number of feature channels \NFeatureChannels.}
From Table~\ref{table:cpmvsres}, increasing it (to 128) for \NStages~$=2$ improves much the performance.
Our \RPM now outperforms the baseline configuration CPM-2S,
especially when considering only the upper body, and is still smaller and slightly faster.
However, from Fig.~\ref{fig:recall-prec-cmp}b, we can notice that the precision-recall curves are not that different between $\NFeatureChannels=64$ and $\NFeatureChannels=128$,
somehow mitigating the above conclusion.

\myparit{Number of recursive stages \NStages.}
Table~\ref{table:cpmvsres} shows that the results saturate when increasing it beyond \NStages~$=2$ (\ie for \NStages~$=3$).

%% In our experiments we experienced increments in performance, specially for recall, by incrementing the number of detector blocks from \NStages~$=1$ to \NStages~$=2$.
%% Nevertheless, we found no benefit to include a third block . Rather, we observed a decrease of $1.65\%$ in precision and of $0.43\%$ in recall.

\myparit{Inpainting preprocessing.}
We found this step to be important to improve the performance of the different tested models.
This is particularly true for our \RPM model, as shown by the curves in Figure~\ref{fig:recall-prec-cmp}(c).
%
%Particularly, the average recall for \TheRPM is improved by $15.74\%$ whereas for CPM-2S is only of $7.08\%$.
We believe that one explanation is that, since the training and test images (synthetic and real) come from different distributions,
this preprocessing removes some of these differences by eliminating  noisy details and discontinuities that typically appear in real depth images.
Figure~\ref{fig:output} shows a qualitative comparison of applying inpainting as preprocessing step.
The figure shows typical multi-person HRI scenarios where person occlusion and partial observations are commonly observed.

\begin{figure}[tb]
\centering
\includegraphics[width=0.49\linewidth]{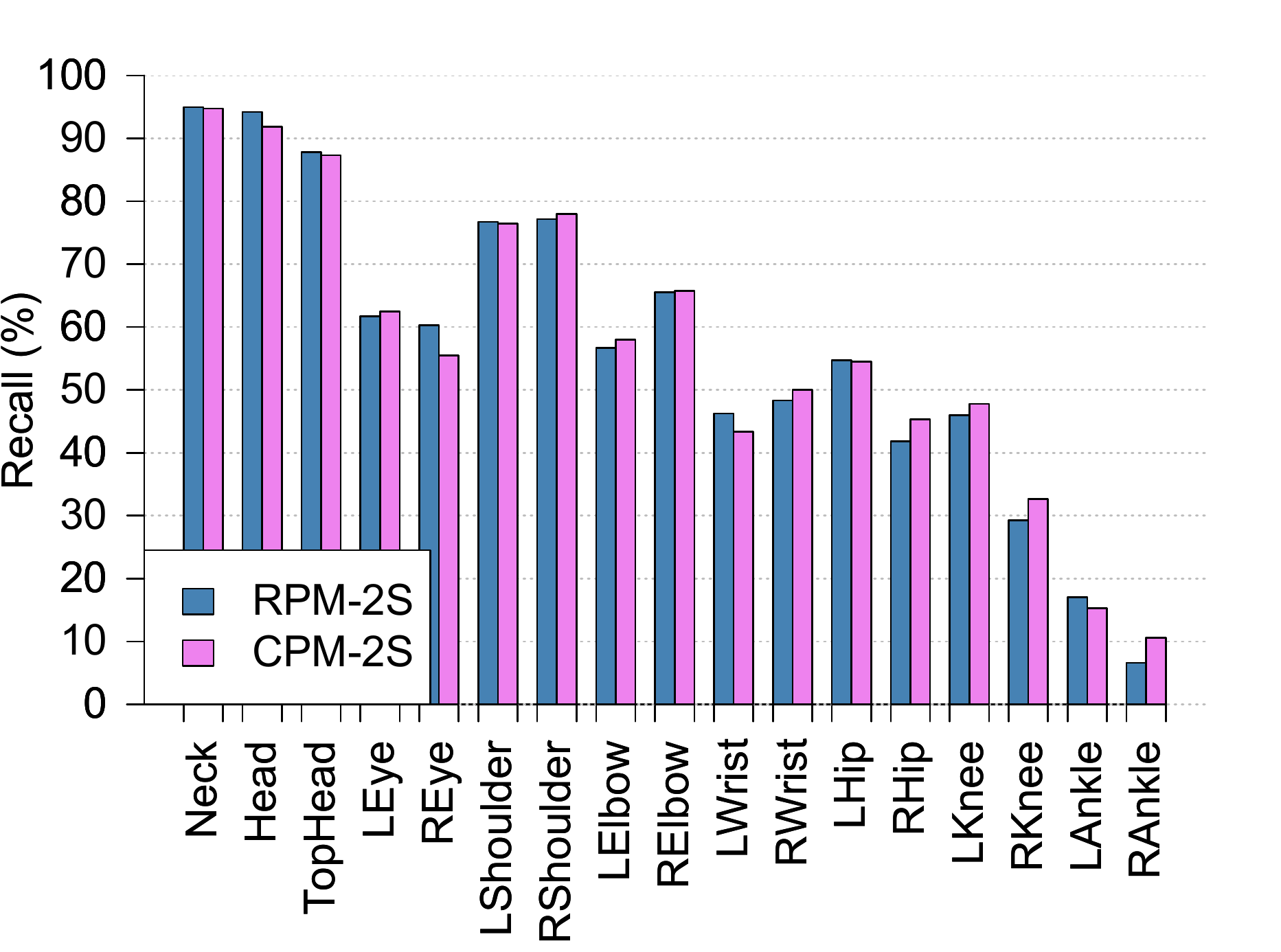}
\includegraphics[width=0.49\linewidth]{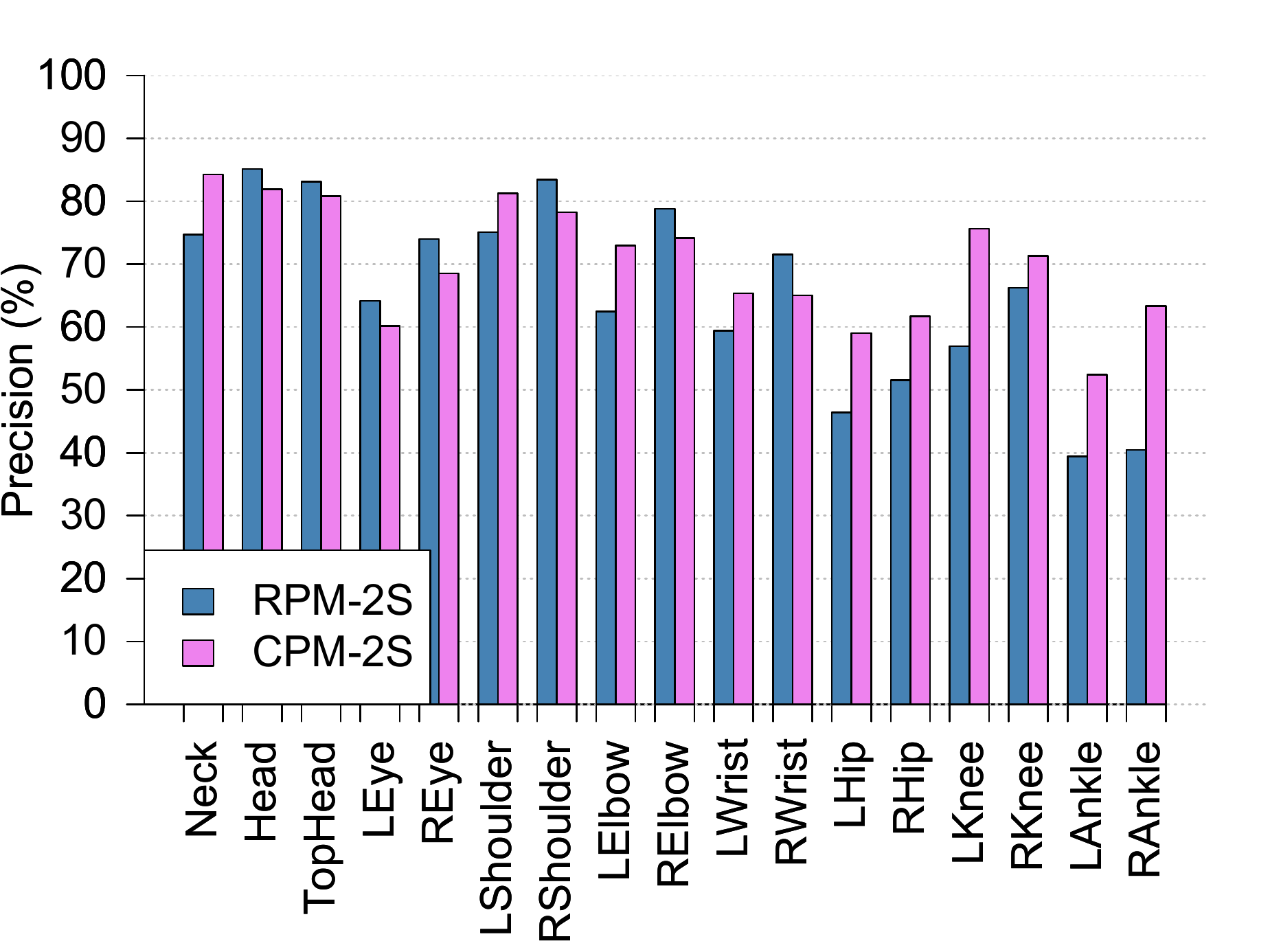}
\vspace{-0.5cm}
\caption{Recall (left) and precision (right) per body landmark for the proposed RPM-2S and the baseline CPM-2S.}
\label{fig:cmp-2S}
\end{figure}

\myparit{Performance per body landmark.}
They are reported in Figure~\ref{fig:cmp-2S} for our  \TheRPM model and CPM-2S.
Both models show similar recall for the different parts of the skeleton. 
As for precision, the difference is more notorious in body parts of the lower body.
Note that in our testing data, these body parts are the most affected by noise,
appearing less well defined and even mixed with the background after the preprocessing.

%%%%%%%%%%%%%%%%%%%%%%%%%%%%%%%%%%%%%%%%%%%%%%%%%%%%%%%%%%%%%%%%%%%%%%%%%%%%%%
\section{CONCLUSIONS}
\label{sec:conclusions}

This paper has investigated the use of depth images and CNNs to perform fast and reliable human pose estimation.
Specifically, we investigated and proposed a real-time neural network architecture with fast forward pass that can be easily deployed
in Human-Robot Interaction applications.
We also created the \OurDataset dataset comprising a large amount of synthetic images of human-bodies of various shapes and
poses, along with real images. This dataset will made publicly available.
Our experiments and speed-accuracy trade-off analysis show  that on depth images,
the  smaller CNN architectures we propose achieve similar performance results as larger versions
with a much less expensive computational cost.

Our study opens the way to further research.
One limitation remains the differences that synthetic depth images exhibit with real ones.
While the inpainting preprocessing mitigates this issue, domain adaptation techniques might be more appropriate
at bridging the existing gap between the data distributions and transfer realistic details to synthetic images.

%                                   % the textheight of the last page by a suitable amount.
%                                   % This command does not take effect until the next page
%                                   % so it should come on the page before the last. Make
%                                   % sure that you do not shorten the textheight too much.

%%%%%%%%%%%%%%%%%%%%%%%%%%%%%%%%%%%%%%%%%%%%%%%%%%%%%%%%%%%%%%%%%%%%%%%%%%%%%%%%

%%%%%%%%%%%%%%%%%%%%%%%%%%%%%%%%%%%%%%%%%%%%%%%%%%%%%%%%%%%%%%%%%%%%%%%%%%%%%%%%

%%%%%%%%%%%%%%%%%%%%%%%%%%%%%%%%%%%%%%%%%%%%%%%%%%%%%%%%%%%%%%%%%%%%%%%%%%%%%%%%
%\section*{APPENDIX}

\section*{ACKNOWLEDGMENTS}

This work was supported by the European Union under the EU Horizon 2020 Research
and Innovation Action MuMMER (MultiModal Mall Entertainment Robot), project ID 688147,
as well as the Mexican National Council for Science and Tecnology (CONACYT) under the PhD scholarships programme.

\bibliographystyle{plain}
\bibliography{bliblio}

%\begin{thebibliography}{99}
%\bibliography{bliblio}
%\end{thebibliography}

\end{document}